\documentclass[10pt,twocolumn,letterpaper]{article}

\usepackage[pagenumbers]{cvpr} 

\usepackage[dvipsnames]{xcolor}

\usepackage{bm}
\usepackage{amsfonts}
\usepackage{multirow}
\usepackage{makecell}
\usepackage{microtype}
\usepackage[accsupp]{axessibility}

\newcommand{\tbf}[1]{\textbf{#1}}
\newcommand{\tul}[1]{\underline{#1}}
\newcommand{\tworow}[2]{\begin{tabular}[c]{@{}c@{}}#1\vspace{-2pt}\\#2\end{tabular}}

\newcommand{\psp}{\kern0.2ex}
\newcommand{\nsp}{\kern-0.1ex}
\newcommand{\pslp}{\psp/\psp}
\newcommand{\LBMTP}{LMTraj\xspace}
\newcommand{\LBMTPZ}{LMTraj\smash{\raisebox{0.2ex}{-}}\nsp ZERO\xspace}
\newcommand{\LBMTPS}{LMTraj\smash{\raisebox{0.2ex}{-}}\nsp SUP\xspace}

\newcommand{\slcb}{\nsp\smash{\raisebox{0.05ex}{\small\{}}\nsp}
\newcommand{\srcb}{\nsp\smash{\raisebox{0.05ex}{\small\}}}\nsp}
\newcommand{\ws}{\nsp\smash{\raisebox{-0.5ex}{\textvisiblespace}}\nsp}

\definecolor{cvprblue}{rgb}{0.21,0.49,0.74}
\usepackage[pagebackref,breaklinks,colorlinks,citecolor=cvprblue]{hyperref}

\def\paperID{1350} 
\def\confName{CVPR}
\def\confYear{2024}

\title{Can Language Beat Numerical Regression?\\Language-Based Multimodal Trajectory Prediction}

\author{Inhwan Bae\textsuperscript{\rm 1}, Junoh Lee\textsuperscript{\rm 2} and Hae-Gon Jeon\textsuperscript{\rm 1,2}\thanks{Corresponding author}\\
\textsuperscript{\rm 1}AI Graduate School, \textsuperscript{\rm 2}School of Electrical Engineering and Computer Science\\
Gwangju Institute of Science and Technology, Gwangju, South Korea\\
{\tt\small \{inhwanbae, juno\}@gm.gist.ac.kr, haegonj@gist.ac.kr}
}

\begin{document}
\maketitle

{%
\begin{abstract}
\vspace{-3mm}
Language models have demonstrated impressive ability in context understanding and generative performance. Inspired by the recent success of language foundation models, in this paper, we propose \LBMTP (Language-based Multimodal Trajectory predictor), which recasts the trajectory prediction task into a sort of question-answering problem. Departing from traditional numerical regression models, which treat the trajectory coordinate sequence as continuous signals, we consider them as discrete signals like text prompts. Specially, we first transform an input space for the trajectory coordinate into the natural language space. Here, the entire time-series trajectories of pedestrians are converted into a text prompt, and scene images are described as text information through image captioning. The transformed numerical and image data are then wrapped into the question-answering template for use in a language model. Next, to guide the language model in understanding and reasoning high-level knowledge, such as scene context and social relationships between pedestrians, we introduce an auxiliary multi-task question and answering.\,We\,then\,train\,a\,numerical tokenizer with the prompt data.\,We\,encourage\,the tokenizer to separate the integer and decimal parts well, and leverage it to capture correlations between the consecutive numbers in the language model. Lastly, we train the language model using the numerical tokenizer and all of the question-answer prompts. Here, we propose a beam-search-based most-likely prediction and a temperature-based multimodal prediction to implement both deterministic and stochastic inferences. Applying\,our\,\LBMTP,\,we\,show that the language-based model can be a powerful pedestrian trajectory predictor, and outperforms existing numerical-based predictor methods. Extensive experiments show that our \LBMTP can successfully understand social relationships and accurately extrapolate the multimodal futures on the public pedestrian trajectory prediction benchmark. Code is publicly available at \url{https://github.com/inhwanbae/LMTrajectory}.
\end{abstract}
\vspace{-10mm}}

\begin{figure}[t]
\vspace{1mm}
\centering
\includegraphics[width=\linewidth,trim={0 0 0 32mm},clip]{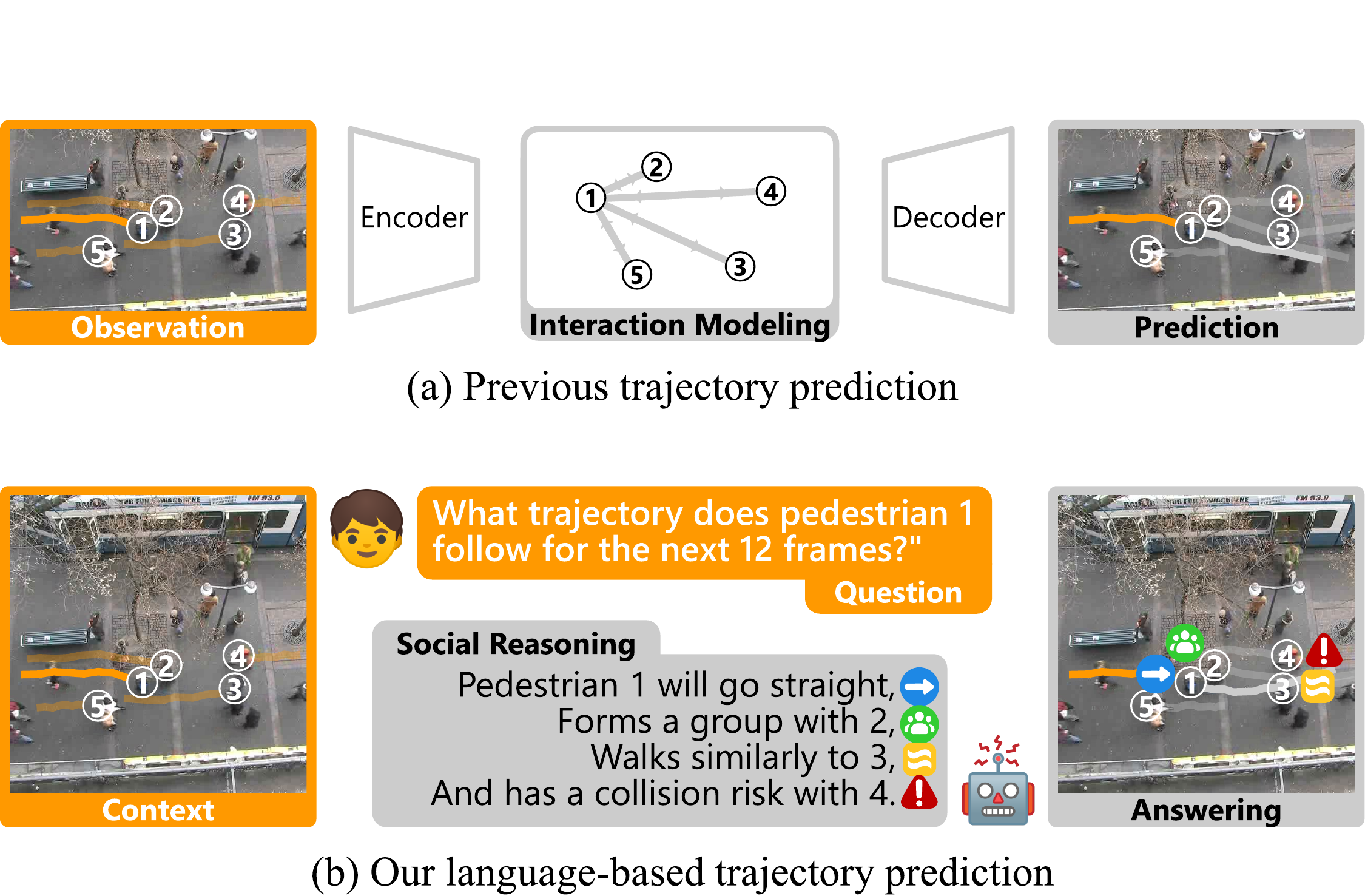}
\vspace{-6mm}
\caption{\textbf{Traditional vs. Our language-based trajectory prediction, \LBMTP}. Given each observation data, (a) traditional predictors directly use the numerical values; (b) the proposed method converts the raw trajectory data to the linguistic prompt, and then captures reasoning social relations to predict a socially acceptable future with the question-answering template.}
\label{fig:teaser}
\vspace{3mm}
\end{figure}

\section{Introduction}
Forecasting pedestrian trajectories in crowded environments is essential for path planning, social robots and autonomous maneuvering systems. The mainstream models used for this task take the position of pedestrians in world-coordinates as input, and infer their possible future paths by regressing a set of coordinate sequences~\cite{alahi2016social,gupta2018social,li2019conditional,fernando2018gdgan,shi2020multimodal,kosaraju2019social,sun2020reciprocal,salzmann2020trajectron++,mohamed2020social,Shi2021sgcn,yu2020spatio,li2020Evolvegraph,mangalam2020pecnet,bae2022npsn,bae2021dmrgcn,liu2021causal,xu2022remember}. Capturing social relations between pedestrians, based on their distance and motion similarity, has resulted in impressive performance improvements~\cite{velivckovic2018graph,vaswani2017attention,mohamed2020social,yu2020spatio,yuan2021agent,Shi2021sgcn}. 

Meanwhile, recent advances in language models have demonstrated their ability to provide context understanding and conditional generation across a spectrum of tasks~\cite{liu2021kg, gong2020tablegpt, ribeiro2021structural}. The language models also offer accurate results when solving mathematical problems~\cite{shridhar2022automatic, zong2023solving}. This is because the language models can provide higher-level connotations~\cite{xue2022leveraging,xue2022prompt,kuo2022trajectory}, and benefit from tokenizers~\cite{kudo2018sentencepiece} and extensive knowledge embedded in large pretrained models~\cite{alec2021clip,radford2019gpt2}. The beauty of the language models is to account for social reasoning~\cite{turiel1983development} beyond the physics-based interactions. As a result, we can intuitively expect improvement in interaction modeling when a language model is introduced in trajectory prediction. However, there are remaining challenges before they can be practically applied: (1) since it is trained on text data, the text tokenizer often does not work on numerical data; (2) it does not consider numerical data with decimal precision; (3) it does not attempt to extrapolate the time-series data using the numerical data itself.

In this paper, we investigate the feasibility of using natural language processing (NLP) to infer the future trajectories of pedestrians. We attempt to bridge the gap between traditional trajectory predictors and the capability of contemporary language models, offering a holistic solution for forecasting in crowded scenarios. Here, we introduce the Language-based Multimodal Trajectory predictor (\LBMTP), which reevaluates language models from their foundational levels for numerical forecasting, as illustrated in~\cref{fig:teaser}. Our \LBMTP consists of four steps:
(1) We convert the raw trajectory coordinates and scene images into textual prompts. The raw coordinates are transformed into a set of decimal notations, and the images are converted into natural language through an image captioning model. Both prompts are then integrated into the question-answering (QA) template as context information.
(2) We introduce supplementary tasks to push the model to learn a higher level of context understanding. Auxiliary questions about the number of group members and collision occurrences drive the model to consider social relationships when forecasting future trajectories. 
(3) We conduct an in-depth analysis of tokenizers, which have been largely overlooked by existing models. Our numerical tokenizer, optimized using the trajectory prompt, splits text and numbers clearly so that the model can learn correlations between sequential natures.
(4) We enable the language model to infer future trajectories in both deterministic and stochastic manners. To generate the most likely and multimodal trajectories, we incorporate beam-search and temperature-tuning techniques.

Lastly, we evaluate the language model as a numerical regressor through both \textit{zero-shot} and \textit{supervised} approaches. We perform a zero-shot evaluation using prompt engineering on the two language foundation models. To take full advantage of the language model, we integrate all the proposed components into our \LBMTP model.
By effectively incorporating the proposed methods, our model achieves state-of-the-art results using a variety of public pedestrian trajectory prediction benchmarks, which are commonly regarded as the area of numerical regressors.

\section{Related Works}
\subsection{Pedestrian Trajectory Prediction}
Beginning with physics-based mathematical formulation methods~\cite{helbing1995social,pellegrini2009you,mehram2009socialforcemodel,yamaguchi2011you}, trajectory forecasting has significantly improved under the numerical-based prediction paradigm. Following advances in convolutional neural networks (CNNs) and recurrent neural networks (RNNs), trajectory prediction become capable of inferring socially-acceptable paths using social interactions and motion modeling. One pioneering work is Social-LSTM~\cite{alahi2016social}, which recurrently predicts future coordinates using a long short-term memory (LSTM), while the social interaction between neighboring agents is modeled by aggregating hidden states via a pooling mechanism. Employing methods such as attention mechanisms~\cite{vemula2018social,ivanovic2019trajectron,fernando2018soft,salzmann2020trajectron++}, graph convolutional networks (GCNs)~\cite{kipf2016semi,mohamed2020social,sun2020rsbg,bae2021dmrgcn}, graph attention networks (GATs)~\cite{velivckovic2018graph,huang2019stgat,kosaraju2019social,liang2020garden,liang2020simaug,Shi2021sgcn,bae2022npsn}, or transformers \cite{yu2020spatio,yuan2021agent,gu2022mid,monti2022stt,bae2022gpgraph,wen2022socialode,wong2022v2net,shi2023trajectory,pourkeshavarz2023learn,chen2023traj,zhu2023biff,choi2023r} allows us to directly model mutual influences among agents. Plus, additional environmental information can lead to better prediction results~\cite{varshneya2017,xue2018sslstm,manh2018scene,sadeghian2019sophie,liang2019peeking,kosaraju2019social,sun2020reciprocal,dendorfer2020goalgan,dendorfer2021mggan,zhao2019matf,tao2020dynamic,sun2020rsbg,marchetti2020mantra,marchetti2020multiple,deo2020trajectory,shafiee2021Introvert,mangalam2021ynet,yue2022nsf}. Subsequent works take either recurrent \cite{alahi2016social,gupta2018social,bisagno2018group,pfeiffer2018,zhang2019srlstm,xu2020cflstm,salzmann2020trajectron++,ma2020autotrajectory,zhao2021experttraj,chen2021disdis,kothari2021interpretable,liu2021snce,sun2022human,lee2022musevae,gu2022mid,marchetti2022smemo,navarro2022social,xu2023uncovering,chen2023unsupervised,wang2023fend,dong2023sparse,maeda2023fast} or simultaneous approaches \cite{mohamed2020social,hug2020bezier,bae2021dmrgcn,Shi2021sgcn,li2021stcnet,pang2021lbebm,shi2022social,bae2023graphtern,bae2023eigentrajectory,aydemir2023adapt} to extrapolate the future trajectories. Recent works combine probabilistic inferences with the bivariate Gaussian distribution \cite{alahi2016social,rehder2018pedestrian,chai2019multipath,mohamed2020social,shi2020multimodal,yu2020spatio,li2020Evolvegraph,shi2021socialdpf,bae2021dmrgcn,yao2021bitrap,Shi2021sgcn,xu2022tgnn,mohamed2022socialimplicit}, Generative Adversarial Network (GAN)~\cite{gupta2018social,sadeghian2019sophie,kosaraju2019social,zhao2019matf,sun2020reciprocal,li2019idl,liang2021tpnms,dendorfer2021mggan,huang2019stgat,sun2023stimulus}, Conditional Variational AutoEncoder (CVAE)~\cite{lee2017desire,li2019conditional,ivanovic2019trajectron,bhattacharyya2020conditional,salzmann2020trajectron++,mangalam2020pecnet,zhao2020tnt,chen2021disdis,sun2021pccsnet,lee2022musevae,wang2022stepwise,xu2022groupnet,xu2022socialvae} and diffusion \cite{gu2022mid,mao2023leapfrog,jiang2023motiondiffuser,rempe2023trace} for multi-modal trajectory generation.

Departing from the mainstream methods, works in~\cite{choi2019looking,mangalam2021ynet} predict heatmaps at the pixel level in images for possible future paths. Like the classification task, some works \cite{liang2019peeking,pajouheshgar2018back,deo2020trajectory} have output classified positions on a discretized (Manhattan) grid. Unfortunately, they reach a limit because the trajectory prediction task requires forecasting accurate pathways based on social norms.

\subsection{Language-Based Reasoning and Prediction}
Transformer architectures and their training schemes have led to the notable development of language foundation models in the NLP field. In particular, BERT~\cite{kenton2019bert} employs a masked language modeling (MLM), which randomly masks a certain percentage of words and trains the model to predict them. GPT-2~\cite{radford2019gpt2} uses a causal language modeling (CLM), an autoregressive method for predicting the next token. T5~\cite{raffel2020t5} involves sequence-to-sequence (Seq2Seq) modeling, using an encoder-decoder architecture to generate the output sequence. These unique models stand out in various generative tasks, including machine translation~\cite{cai2021nmt, yang2020towards, xu2021improvingmultilingual}, text generation~\cite{liu2021kg, gong2020tablegpt, ribeiro2021structural}, and question-answering~\cite{Dong2019Unified, bao2020unilmv2, qi2020prophetnet}.

Beyond the NLP field, language foundational models have also exhibited superior performance in vision-language tasks and solving mathematical problems. This includes classification~\cite{alec2021clip,zhou2021learning}, generation~\cite{ding2021cogview, zhou2022ttigeneration}, and problem-solving~\cite{shridhar2022automatic, zong2023solving}. These works explore the application of foundational language models, with the goal of extending the scope of the pre-training/fine-tuning paradigm.

Most recently, there have been attempts to incorporate language priors into time-series forecasting~\cite{spithourakis2018numeracy,berg2020empirical,na2023spu}. For instance, ForecastQA~\cite{jin2020forecastqa} proposes a QA benchmark with timestamp constraints to verify its forecasting ability regarding future events. Xue \etal~\cite{xue2022leveraging} study mobility prediction, inferring how people move in cities. Inspired by chatbot applications, PromptCast~\cite{xue2022prompt} has made predictions on weather temperature, energy consumption, and customer flow. The most relevant work to ours~\cite{kuo2022trajectory} uses linguistic intermediate representations for trajectory prediction, solving action-related reasoning through language priors. However, they cannot fully take advantage of the linguistic model, in that they inherently use pre-trained tokenizers learned from text data. In particular, their approach is not suitable for trajectory prediction tasks because of the inconsistent analysis of numerical data. Furthermore, when dealing with coordinate sequences, existing numerical regressors are directly utilized as auxiliary modules to language models, inhibiting a higher level of understanding like social interactions.

\section{Methodology}\label{sec:method}
Our approach shifts the paradigm from conventional trajectory prediction to a prompt-based perspective. We recast the trajectory prediction task in a sentence-to-sentence manner, which uses the numerical input and output as a prompt and applies a language model for the purpose of numerical forecasting. In this work, we propose a language-based trajectory prediction framework, \LBMTP, consisting of \LBMTPZ and \LBMTPS, using both zero-shot and supervised approaches, respectively.

We start with a numerical definition of the trajectory forecasting task in~\cref{sec:method_problem_definition}. We then describe our considerations in converting the numerical trajectories and images into text prompts and designing prompt templates to obtain desirable responses from language models for both \LBMTPZ and \LBMTPS in~\cref{sec:method_prompt_conversion}. Using these text prompts, we obtain the best performance with the language model-based trajectory predictor, \LBMTPS, as described in~\cref{sec:method_domain_shift}. In~\cref{sec:method_forecasting}, we introduce how to build the language models, whose implementation details can be found in~\cref{sec:method_implementation}.

\subsection{Problem Definition}\label{sec:method_problem_definition}
The problem of trajectory prediction involves forecasting the time-series future coordinates of each agent from their historical coordinate sequences. This task can be regarded as a sequence-to-sequence problem. Formally, given a scene image $\mathcal{I}$ and a past observation trajectory with length $T_{\textit{obs}}$, it can be denoted as $\mathcal{S}_{n,\textit{\,obs}}\!=\!\{ (x_n^t, y_n^t)\!\in\!\mathbb{R}^2\,|\,t\!\in\![1, ..., T_{\textit{obs}}] \}$, where $(x_n^t, y_n^t)$ is the 2D coordinate of a specific pedestrian $n$ at time $t$. In the same way, a ground truth future trajectory for the prediction length $T_{\textit{pred}}$ can be written as $\mathcal{S}_{n,\textit{\,pred}}\!=\!\{ (x_n^t, y_n^t)\!\in\!\mathbb{R}^2\,|\,t\!\in\![T_{\textit{obs}}\!+\!1, ..., T_{\textit{obs}}\!+\!T_{\textit{pred}}] \}$. The prediction model takes both $\mathcal{S}_{\textit{obs}}$ and $\mathcal{I}$ as input. It either predicts one most-likely path $\widehat{\mathcal{S}}_{\textit{pred}}$ or generates $K$ possible multi-modal future trajectories $\widehat{\mathcal{S}}_{\textit{pred}}^{\,k}$, which are called deterministic and stochastic predictions, respectively.

\subsection{Data Space Conversion to Prompt}\label{sec:method_prompt_conversion}
To make predictions using a language model, we first need to convert the raw data into text prompts. The most common data used in trajectory prediction is a numerical coordinate sequence and top-down view images of a scene. In this section, we start by transforming the pedestrian trajectory and environmental data. The converted data are then aggregated into linguistic sentences using a QA template for the input and output of the language model.

\vspace{0.5mm}\noindent\textbf{Converting trajectory coordinates into the prompt.}\quad
We convert the entire float-type coordinate value to a text string with decimal representation. Compared to the binary numerical system, which is commonly used for network input, decimal representation is more compatible with natural language. In this process, we round the continuous values to discrete values with two decimal places for the word coordinate system in order to efficiently use the prompt. We leave it as an integer value if the trajectory is in the pixel coordinate system. Second, to represent the sequence of 2D coordinates, we concatenate the $x_n^t$ and $y_n^t$ coordinates using a comma separator and round bracket, and combine the time-series coordinates $\{ (x_n^t, y_n^t)\,|\,t \}$ using the square bracket, as shown in~\cref{tab:method_template}. By using the different bracket symbols, it becomes easier to parse the spatio-temporal information. With this trick, we transform both the history and future trajectories $\mathcal{S}_{n,\textit{\,obs}}$, $\mathcal{S}_{n,\textit{\,pred}}$ into text prompts $\mathcal{P}_{\mathcal{S}_{n,\textit{\,obs}}}$, $\mathcal{P}_{\mathcal{S}_{n,\textit{\,pred}}}$, and repeat the process for all $N$ pedestrians in the scene.

\vspace{0.5mm}\noindent\textbf{Converting image data into the prompt.}\quad
We convert the scene image $\mathcal{I}$ into prompts $\mathcal{P}_\mathcal{I}$ as well. Inspired by image captioning, we employ the BLIP-2 model~\cite{li2023blip2}, trained on ImageNet~\cite{ilsvrc15imagenet}, to extract text descriptions that depict the agent-moving scene. Taking the image description prompt as input, the model is able to learn various environmental details, such as the placement of buildings and vehicles, the density of people, and the flow of pedestrians. This helps the model to determine moving speeds and behavior patterns, similar to a traditional map encoding using pretrained segmentation models~\cite{mangalam2021ynet}.

\vspace{0.5mm}\noindent\textbf{Converting predictions into the prompt.}\quad
Next, the numerical coordinate prompt and the scene description prompt are preprocessed before being fed into \LBMTP. For trajectory prediction with a language model, we need to make it suitable for the NLP task. Note that the QA task gives context information to a language model and asks questions to ensure the correct answers. We introduce a question-answering template $\mathcal{T}_\textit{forecast} \!=\! \{\mathcal{P}_\mathcal{C}, \mathcal{P}_\mathcal{Q}, \mathcal{P}_\mathcal{\!A}\}$ for trajectory forecasting. We provide the history coordinates of all agents in a scene as context $\mathcal{P}_\mathcal{C}$ and ask the model to predict the future trajectory for a specific pedestrian $n$ using $\mathcal{P}_\mathcal{Q}$. The answer we expect is $\mathcal{P}_\mathcal{\!A}$. This template-based description can effectively transform the data into text~\cite{xue2022translating}.

\begin{table}
\centering
\resizebox{\linewidth}{!}{%
\textls[-15]{
\begin{tabular}{@{~~}c@{~}c@{~}c@{~~}l@{}}
\toprule
Prompt & Type & Field & Template \\ \midrule
$\mathcal{P}_{\mathcal{S}_n,\textit{\,obs}}$       & -                       & -                         & ``[(\slcb$x_n^1$\srcb,\ws\slcb$y_n^1$\srcb),\ws(\slcb$x_n^2$\srcb,\ws\slcb$y_n^2$\srcb),\ws$...$,\ws(\slcb$x_n^{T_{\textit{obs}}}$\srcb,\ws\slcb$y_n^{T_{\textit{obs}}}$\srcb)]''                                          \\
\multirow{2}{*}{$\mathcal{T}_{\mathcal{S}_n,\textit{\,obs}}$} & \multirow{2}{*}{-} & \multirow{2}{*}{-}  & ``Pedestrian\ws\slcb$n$\srcb\ws moved\ws along\ws the\ws trajectory\ws\slcb$\mathcal{P}_{\mathcal{S}_n,\textit{\,obs}}$\srcb\ws for\ws                                                                       \vspace{-2pt} \\
                                                   &                         &                           & ~~\slcb$T_{\textit{obs}}$\srcb\ws frames.''                                                                                                                                                                                \\ \midrule
\multirow{5}{*}{\!\!\!\!$\mathcal{T}_{\textit{forecast}}$\!\!} & \multirow{3}{*}{Input}  & \multirow{2}{*}{Question} & ``What\ws trajectory\ws does\ws pedestrian\ws\slcb$n$\srcb\ws follow\ws for\ws the\ws next\ws\slcb$T_{\textit{obs}}$\srcb\ws                                                                                 \vspace{-2pt} \\
                                                   &                         &                           & ~~frames?''                                                                                                                                                                                                                \\
                                                   &                         & Context                   & ``\slcb$\mathcal{P}_\mathcal{I}$\srcb\ws\slcb$\mathcal{T}_{\mathcal{S}_1,\textit{\,obs}}$\srcb\ws\slcb$\mathcal{T}_{\mathcal{S}_2,\textit{\,obs}}$\srcb\ws$...$\ws\slcb$\mathcal{T}_{\mathcal{S}_N,\textit{\,obs}}$\srcb'' \\
                                                   & \multirow{2}{*}{Output} & \multirow{2}{*}{Answer}   & ``Pedestrian\ws\slcb$n$\srcb\ws will\ws move\ws along\ws the\ws trajectory\ws\slcb$\mathcal{S}_{n,\textit{\,pred}}$\srcb\ws for\ws the\ws                                                                    \vspace{-2pt} \\
                                                   &                         &                           & ~~next\ws\slcb$T_{\textit{pred}}$\srcb\ws frames.''                                                                                                                                                                        \\ \cmidrule(lr){1-4}
\multirow{5}{*}{$\mathcal{T}_\textit{dest}$}       & \multirow{3}{*}{Input}  & \multirow{2}{*}{Question} & ``At\ws which\ws coordinates\ws does\ws pedestrian\ws\slcb$n$\srcb\ws arrive\ws after\ws the\ws next\ws                                                                                                      \vspace{-2pt} \\
                                                   &                         &                           & ~~\slcb$T_{\textit{pred}}$\srcb\ws frames?''                                                                                                                                                                                \\
                                                   &                         & Context                   & ``\slcb$\mathcal{P}_\mathcal{I}$\srcb\ws\slcb$\mathcal{T}_{\mathcal{S}_1,\textit{\,obs}}$\srcb\ws\slcb$\mathcal{T}_{\mathcal{S}_2,\textit{\,obs}}$\srcb\ws$...$\ws\slcb$\mathcal{T}_{\mathcal{S}_N,\textit{\,obs}}$\srcb'' \\
                                                   & \multirow{2}{*}{Output} & \multirow{2}{*}{Answer}   & ``Pedestrian\ws\slcb$n$\srcb\ws will\ws arrive\ws at\ws coordinate                                                                                                                                           \vspace{-2pt} \\
                                                   &                         &                           & ~~(\slcb$x_n^{\smash{T_{\textit{obs}}+T_{\textit{pred}}}}$\srcb,\ws\slcb$y_n^{\smash{T_{\textit{obs}}+T_{\textit{pred}}}}$\srcb)\ws after\ws the\ws next\ws\slcb$T_{\textit{pred}}$\srcb\ws frames.''                      \\ \cmidrule(lr){1-4}
\multirow{4}{*}{$\mathcal{T}_\textit{dir}$}        & \multirow{2}{*}{Input}  & Question                  & ``In\ws which\ws direction\ws will\ws pedestrian\ws\slcb$n$\srcb\ws move\ws in\ws the\ws future?''                                                                                                                         \\
                                                   &                         & Context                   & ``\slcb$\mathcal{P}_\mathcal{I}$\srcb\ws\slcb$\mathcal{T}_{\mathcal{S}_1,\textit{\,obs}}$\srcb\ws\slcb$\mathcal{T}_{\mathcal{S}_2,\textit{\,obs}}$\srcb\ws$...$\ws\slcb$\mathcal{T}_{\mathcal{S}_N,\textit{\,obs}}$\srcb'' \\
                                                   & \multirow{2}{*}{Output} & \multirow{2}{*}{Answer}   & ``Pedestrian\ws\slcb$n$\srcb\ws will                                                                                                                                                                         \vspace{-2pt} \\
                                                   &                         &                           & ~~\slcb$\text{move\ws forward} \psp|\psp \text{move\ws backward} \psp|\psp \text{move\ws left} \psp|\psp \text{move\ws right} \psp|\psp \text{stop}$\srcb.''                                                               \\ \cmidrule(lr){1-4}
\multirow{4}{*}{$\mathcal{T}_\textit{mimic}$}      & \multirow{2}{*}{Input}  & Question                  & ``Which\ws pedestrian\ws seems\ws to\ws walk\ws similarly\ws to\ws pedestrian\ws\slcb$n$\srcb?''                                                                                                                           \\
                                                   &                         & Context                   & ``\slcb$\mathcal{P}_\mathcal{I}$\srcb\ws\slcb$\mathcal{T}_{\mathcal{S}_1,\textit{\,obs}}$\srcb\ws\slcb$\mathcal{T}_{\mathcal{S}_2,\textit{\,obs}}$\srcb\ws$...$\ws\slcb$\mathcal{T}_{\mathcal{S}_N,\textit{\,obs}}$\srcb'' \\
                                                   & \multirow{2}{*}{Output} & \multirow{2}{*}{Answer}   & Case 1: ``Pedestrian\ws\slcb$n$\srcb\ws walks\ws similarly\ws to\ws pedestrian\ws\slcb$k$\srcb.''                                                                                                            \vspace{-2pt} \\
                                                   &                         &                           & Case 2: ``Pedestrian\ws\slcb$n$\srcb\ws will\ws walk\ws alone.''                                                                                                                                                           \\ \cmidrule(lr){1-4}
\multirow{4}{*}{$\mathcal{T}_\textit{group}$}      & \multirow{2}{*}{Input}  & Question                  & ``With\ws which\ws pedestrians\ws does\ws pedestrian\ws\slcb$n$\srcb\ws form\ws a\ws group?''                                                                                                                              \\
                                                   &                         & Context                   & ``\slcb$\mathcal{P}_\mathcal{I}$\srcb\ws\slcb$\mathcal{T}_{\mathcal{S}_1,\textit{\,obs}}$\srcb\ws\slcb$\mathcal{T}_{\mathcal{S}_2,\textit{\,obs}}$\srcb\ws$...$\ws\slcb$\mathcal{T}_{\mathcal{S}_N,\textit{\,obs}}$\srcb'' \\
                                                   & \multirow{2}{*}{Output} & \multirow{2}{*}{Answer}   & Case 1: ``Pedestrian\ws\slcb$n$\srcb\ws forms\ws a\ws group\ws with\ws pedestrian\ws\slcb$k$\srcb.''                                                                                                         \vspace{-2pt} \\
                                                   &                         &                           & Case 2: ``Pedestrian\ws\slcb$n$\srcb\ws will\ws walk\ws alone.''                                                                                                                                                           \\ \cmidrule(lr){1-4}
\multirow{4}{*}{$\mathcal{T}_\textit{col}$}        & \multirow{2}{*}{Input}  & Question                  & ``With\ws which\ws pedestrian\ws does\ws pedestrian\ws\slcb$n$\srcb\ws have\ws a\ws collision\ws risk?''                                                                                                                   \\
                                                   &                         & Context                   & ``\slcb$\mathcal{P}_\mathcal{I}$\srcb\ws\slcb$\mathcal{T}_{\mathcal{S}_1,\textit{\,obs}}$\srcb\ws\slcb$\mathcal{T}_{\mathcal{S}_2,\textit{\,obs}}$\srcb\ws$...$\ws\slcb$\mathcal{T}_{\mathcal{S}_N,\textit{\,obs}}$\srcb'' \\
                                                   & \multirow{2}{*}{Output} & \multirow{2}{*}{Answer}   & Case 1: ``Pedestrian\ws\slcb$n$\srcb\ws has\ws a\ws collision\ws risk\ws with\ws pedestrian\ws\slcb$k$\srcb.''                                                                                               \vspace{-2pt} \\
                                                   &                         &                           & Case 2: ``Pedestrian\ws\slcb$n$\srcb\ws has\ws no\ws collision\ws risk.''                                                                                                                                                  \\
\bottomrule
\end{tabular}
}}
\vspace{-3mm}
\caption{QA templates to convert raw trajectory data into prompts.}
\label{tab:method_template}
\end{table}

\subsection{Domain Shift to Sentence Generation}\label{sec:method_domain_shift}
After the conversion process, we revisit each component of the conventional NLP pipelines, and introduce a domain adaptation for \LBMTPS.

\vspace{0.5mm}\noindent\textbf{Optimizing the tokenizer for numeric data.}\quad
The first thing that we revisit is the tokenizer. A tokenizer is an essential component that breaks text down into smaller units called tokens, which are used as the preliminary step in conventional NLP models to parse and understand languages~\cite{NIPS2013Distributed,pennington2014glove, kudo2018subword,kudo2018sentencepiece}. 

\begin{figure}[t]
\centering
\includegraphics[width=\linewidth,trim={0 0 0 0},clip]{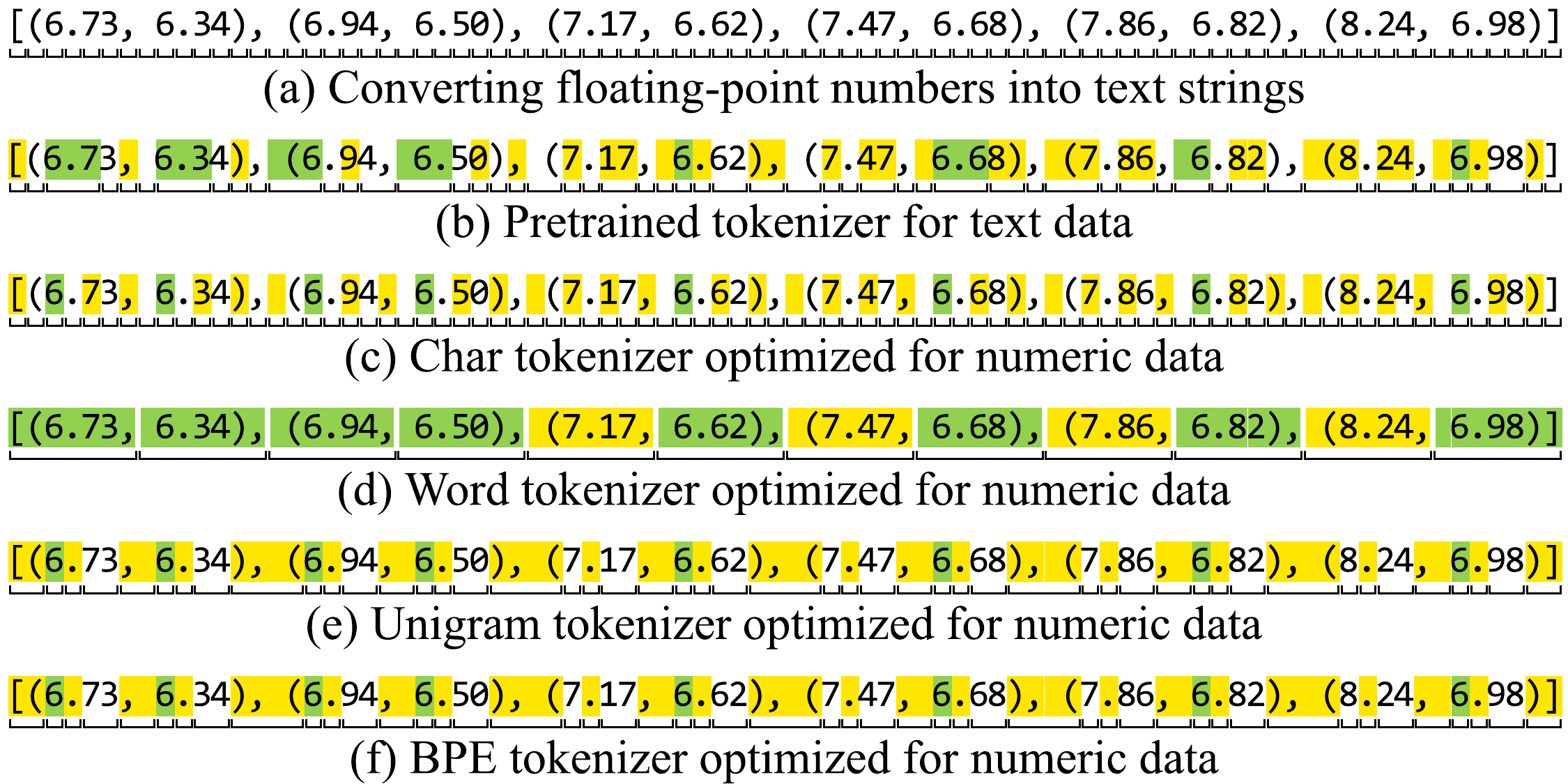}
\vspace{-6.5mm}
\caption{Comparison of the text-pretrained tokenizer and our numeric data-optimized tokenizer. Under brackets with yellow or white highlight colors indicate that the corresponding letters have been tokenized. The green color highlights that the token contains an integer with 6.}
\label{fig:comparison_tokenizer}
\vspace{5mm}
\end{figure}

Following a conventional NLP pipeline~\cite{wu2016google,paulus2018a}, we use a tokenizer to convert the QA prompt into a form that the \LBMTPS can understand. In this step, existing studies directly employ pretrained tokenizers~\cite{NIPS2013Distributed,pennington2014glove}. However, we figure out that when they are optimized for text data, they often fail to properly represent the numerical data. When using this tokenizer, numbers are irregularly split into tokens, and occasionally, special characters like periods and commas are grouped together, as shown in~\cref{fig:comparison_tokenizer}(b). This can disturb the training for consecutiveness and associations between adjacent numbers. 

To address this issue, we train a new tokenizer for the numerical data using our QA prompts $\mathcal{P}_\mathcal{C}$, $\mathcal{P}_\mathcal{Q}$ and $\mathcal{P}_\mathcal{\!A}$ consisting of numerical coordinates and image description prompts. 
As demonstrated in~\cref{fig:comparison_tokenizer}(f), our numerical tokenizer clearly breaks down words, integers and decimal parts well. In addition, because the total number of tokens for the same sentence is reduced by removing the unnecessary splitting problem, \LBMTPS can become lighter and faster.

\vspace{0.5mm}\noindent\textbf{Multi-task training for social relation reasoning.}\quad
In trajectory prediction tasks, the most crucial component is modeling interactions between agents. To enhance the reasoning capacity with social relations, we develop a training scheme for our \LBMTPS. It is a widely known technique which allows language models to achieve high-level knowledge understanding through multi-task learning~\cite{kenton2019bert,raffel2020t5,radford2019gpt2}. While \LBMTPS can learn to perform prediction tasks only with the forecasting QA prompt, we introduce auxiliary tasks to fully take advantage of its understanding and reasoning ability for both scene context and social dynamics. 

The five auxiliary tasks are as follows: destination suggestion, moving direction prediction, similar pattern search, group member prediction, and collision possibility assessment. We implement these synthetic tasks using pseudo labels and QA templates $\mathcal{T}_\textit{dest}$, $\mathcal{T}_\textit{dir}$, $\mathcal{T}_\textit{mimic}$, $\mathcal{T}_\textit{group}$ and $\mathcal{T}_\textit{col}$ in the same way as the forecasting task. \Cref{tab:method_template} lists up the prompt templates, and \LBMTPS yields the six types of outputs. By explicitly teaching various social relations, \LBMTPS can better capture, understand, and use social norms. Among the outputs, \LBMTPS can extract common features for agent motions, and leverage social knowledge (e.g., group walking and collision avoidance) learned from each auxiliary task to enhance the fidelity of the main forecasting task.

\vspace{0.5mm}\noindent\textbf{Generating most-likely and multimodal outputs.}\quad
Lastly, we tune the text generation stage, where \LBMTPS infers the output. In trajectory prediction, it is crucial to generate all possible multiple paths \smash{$\widehat{\mathcal{S}}_{\textit{pred}}^{\,k}$} or the single most likely path \smash{$\widehat{\mathcal{S}}_{\textit{pred}}$}. Numerical regression-based methods predict a deterministic path using encoder-decoder architectures, and extend it to stochastic inferences for diverse multiple paths by introducing a random latent vector as an additional input.

In the same way, to use a language model in this task, it must be able to produce diverse multiple outputs. We assume that if we can effectively leverage the stochasticity of the language model, inherently based on distributional semantics~\cite{cho2014learning}, it can function as a probability-based numerical approach. We handle the stochasticity by introducing a text-generation technique. Using beam search~\cite{freitag2017beam}, \LBMTPS can predict the path $\widehat{\mathcal{P}}_\mathcal{\!A}$ with the highest probability search controlled by a hyperparameter on a depth $d$. On the other hand, the model can generate diverse outputs $\widehat{\mathcal{P}}_\mathcal{\!A}^{\,k}$ by modulating the token probability using a temperature parameter $\tau$ in \LBMTPS. By using these tricks, the language-based model can perform at par with and even potentially replace existing predictor methods.

\begin{table}[t]
\large
\centering
\resizebox{\linewidth}{!}{%
\begin{tabular}{cc@{~~}c@{~~}c@{~~}cc@{~~}cc@{~~}c}
\toprule
\multirow{2}{*}{Tokenizer\vspace{-5pt}} & \multicolumn{4}{c}{Summary} & \multicolumn{2}{c}{Input sentence} & \multicolumn{2}{c}{Output sentence} \\ \cmidrule(lr){2-5} \cmidrule(lr){6-7} \cmidrule(lr){8-9}
           & \#\,Vocab & \#\,Mixed & Clarity & Cover & \#\,Token & Rouge & \#\,Token & Rouge \\ \midrule
Pretrained & 32000      & \tul{504} & \tul{98.43} & 1.00     & 566.25       & 1.00    & 44.47       & 1.00    \\ \cmidrule(lr){1-9}
Char       & \tbf{59}   & \tbf{0}   & \tbf{100}   & 1.00     & 952.80       & 1.00    & 77.48       & 1.00    \\
Word       & 13586      & 13497     & 0.655       & 1.00     & \tbf{142.21} & 1.00    & \tbf{12.10} & 1.00    \\
Unigram    & \tul{1113} & \tbf{0}   & \tbf{100}   & 1.00     & 421.63       & 1.00    & \tul{27.46} & 1.00    \\
BPE        & 1224       & \tbf{0}   & \tbf{100}   & 1.00     & \tul{402.73} & 1.00    & \tul{27.46} & 1.00    \\
\bottomrule
\end{tabular}
}
\vspace{-3mm}
\caption{Evaluation of the tokenizer characteristics. \#\,Vocab: the total number of unique words in the tokenizer, \#\,Mixed: The number of unique entries that contain both characters and numerals, Clarity: Percentage of non-mixed cases in vocab, Cover: Coverage of the tokenizer that can cover all sentences in the dataset, \#\,Token: The average number of tokens per sentence. Rouge: ROUGE-1 score between the original sentences and their reconstructed ones after tokenization.}
\label{tab:result_tokenizer}
\end{table}

\subsection{Forecasting With the Language Model}\label{sec:method_forecasting}
\textls[-3]{Lastly, we incorporate our proposed methods into the trajectory forecasting model. To do this, we adopt two widely-used approaches in computer vision and natural language tasks: (1) conducting zero-shot evaluation through prompt engineering of a pretrained language foundation model, \LBMTPZ and (2) an end-to-end supervision, \LBMTPS.}

\vspace{0.5mm}\noindent\textbf{\LBMTPZ: Zero-shot prediction in the language foundation model.}\quad
Prompt-tuning is a method that fine-tunes a language model, not by retraining it but by optimizing the input prompt that goes into a frozen pre-trained model to produce a desired output~\cite{petroni2019language,liang2021prefix}. The advantage of prompt-tuning is that it allows us to leverage the existing/extensive knowledge embedded within large pre-trained models. 

In this work, we also use pre-trained large language models, GPT-3.5 and GPT-4, not trained for the purpose of trajectory forecasting. Following~\cite{xue2022prompt}, we optimize the input prompt with the following steps: (1) We make an initial forecasting QA prompt $\mathcal{P}_\mathcal{Q}$ to instruct \LBMTPZ on what the desired output should be; (2) The prompts are fed into \LBMTPZ; (3) The outputs $\widehat{\mathcal{P}}_\mathcal{\!A}^{\,k}$ are evaluated by transforming them back into the numerical coordinates $\widehat{\mathcal{S}}_{\textit{pred}}^{\,k}$, ensuring a fair comparison with the conventional metrics. In all the processes, the language model is frozen and is neither trained nor fine-tuned.

\begin{table}[t]
\Large
\centering
\resizebox{\linewidth}{!}{
\begin{tabular}{c@{~~~~~}cccc@{~~~~~~}cc}
\toprule
\multirow{2}{*}{\!\!\!\!Zero-shot\!\!\!\!\vspace{-5pt}} & \multirow{2}{*}{Stop\vspace{-5pt}} & \multirow{2}{*}{Linear\vspace{-5pt}} & \multirow{2}{*}{\tworow{Kalman}{filter}\vspace{-5pt}} & \multirow{2}{*}{\!\!\tworow{AutoTraj-}{ectory\,\cite{ma2020autotrajectory}}\!\!\vspace{-5pt}} & \multicolumn{2}{c}{\tbf{\LBMTPZ}} \\ \cmidrule(lr){6-7}
      &                    &                     &                           &                     & \tbf{\!-GPT-3.5}    & \tbf{\!-GPT-4}            \\ \midrule
ETH   & 2.84\pslp4.82      & 1.00\pslp2.23       & \tul{0.94}\pslp2.13       & N\,/\,A       & 1.07\pslp\tul{1.82} & \tbf{0.80}\pslp\tbf{1.64} \\
HOTEL & 1.15\pslp2.09      & 0.32\pslp0.62       & \tul{0.26}\pslp\tul{0.50} & N\,/\,A       & 0.42\pslp0.65       & \tbf{0.20}\pslp\tbf{0.37} \\
UNIV  & 1.36\pslp2.47      & \tul{0.52}\pslp1.17 & 0.55\pslp1.20             & 0.89\pslp1.45       & 0.56\pslp\tul{0.98} & \tbf{0.37}\pslp\tbf{0.77} \\
ZARA1 & 2.51\pslp4.61      & \tul{0.43}\pslp0.96 & 0.45\pslp0.98             & 0.48\pslp\tul{0.91} & 0.47\pslp\tul{0.91} & \tbf{0.33}\pslp\tbf{0.66} \\
ZARA2 & 1.38\pslp2.53      & \tul{0.33}\pslp0.73 & 0.34\pslp0.75             & 0.50\pslp1.03       & 0.39\pslp\tul{0.71} & \tbf{0.24}\pslp\tbf{0.50} \\ \cmidrule(lr){1-7}
AVG   & 1.85\pslp3.31      & 0.52\pslp1.14       & \tul{0.51}\pslp1.11       & 0.62\pslp1.13       & 0.58\pslp\tul{1.01} & \tbf{0.39}\pslp\tbf{0.79} \\
SDD   & 64.0\pslp116.7\!\! & 18.8\pslp38.0       & \tul{16.6}\pslp33.9       & N\,/\,A       & 17.8\pslp\tul{29.1} & \tbf{10.9}\pslp\tbf{21.0} \\
GCS   & 76.0\pslp138.8\!\! & 18.9\pslp40.7       & \tul{18.3}\pslp\tul{39.4} & N\,/\,A       & 27.7\pslp44.8       & \tbf{12.7}\pslp\tbf{25.5} \\
\bottomrule
\end{tabular}
}
\vspace{-3mm}
\caption{Comparison of \LBMTPZ methods with other zero-shot methods (ADE\pslp FDE, Unit: meter for ETH\pslp UCY and pixel for SDD\pslp GCS). \textbf{Bold}: Best, \underline{Underline}: Second best.}
\label{tab:result_zeroshot}
\end{table}

\vspace{0.5mm}\noindent\textbf{\LBMTPS: Supervision of language-based predictor.}\quad
Next, we evaluate the maximum capacity and performance of the language model through end-to-end training. First of all, we analyze various structures of sentence-to-sentence language models to choose the best model for forecasting. In trajectory prediction, it has been proven that predicting the trajectories through an encoder-decoder architecture is better than using a procedural generation from recurrent models due to the error accumulation issue~\cite{mangalam2020pecnet,bae2023graphtern,mohamed2020social}. Therefore, instead of using a CLM, we choose the Seq2Seq model, an encoder-decoder language model. Similar to the zero-shot predictors, the set of context and question sentences $\{\mathcal{P}_\mathcal{C}, \mathcal{P}_\mathcal{Q}\}$ are given to the network, and the output sentences $\widehat{\mathcal{P}}_\mathcal{\!A}$ are transformed back into numerical data $\widehat{\mathcal{S}}_{\textit{pred}}$. The difference between \LBMTPZ and \LBMTPS lies in the multi-task QA template, whose loss is back-propagated to train \LBMTPS.

\begin{table*}[t]
\Large
\centering
\resizebox{\linewidth}{!}{
\begin{tabular}{c@{~~~~~~}cccccccccccc@{~~~~~~}c}
\toprule
Deterministic & \tworow{Social-}{LSTM\,\cite{alahi2016social}} & \tworow{Social-}{GAN\,\cite{gupta2018social}} & \tworow{SR-LSTM$^\dagger$\!}{\cite{zhang2019srlstm}} & \tworow{STGAT}{\cite{huang2019stgat}} & \tworow{STAR-D$^\dagger$\!}{\cite{yu2020spatio}} & \tworow{Trajectron}{++$^\dagger$\,\cite{salzmann2020trajectron++}} & \tworow{PECNet}{\cite{mangalam2020pecnet}} & MID\,\cite{gu2022mid} & \tworow{GP-Graph}{\cite{bae2022gpgraph}} & \tworow{\!SocialVAE\!}{\cite{xu2022socialvae}} & NPSN\,\cite{bae2022npsn} & \tworow{EigenTraj}{ectory\,\cite{bae2023eigentrajectory}} & \tbf{\LBMTPS} \\ \midrule
ETH   & 1.09 / 2.35 & 1.13 / 2.21 & 1.01 / 1.93 & \tul{0.88} / \tul{1.66} & 0.97 / 2.00 & 1.02 / 2.09 & 1.20 / 2.73 & 1.42 / 2.94 & 0.89 / 1.78 & 0.97 / 1.93 & 0.95 / 2.04 & 0.92 / 2.03 & \tbf{0.65} / \tbf{1.04} \\
HOTEL & 0.79 / 1.76 & 1.01 / 2.18 & 0.35 / 0.72 & 0.56 / 1.15 & 0.32 / 0.73 & 0.33 / 0.63 & 0.68 / 1.51 & 0.64 / 1.30 & 0.47 / 1.03 & 0.40 / 0.78 & 0.32 / \tul{0.57} & \tul{0.29} / \tul{0.57} & \tbf{0.26} / \tbf{0.46} \\
UNIV  & 0.67 / 1.40 & 0.60 / 1.28 & 0.66 / 1.38 & \tbf{0.52} / \tbf{1.13} & 0.56 / 1.25 & \tbf{0.52} / \tul{1.16} & 0.78 / 1.71 & 0.76 / 1.62 & 0.56 / 1.19 & \tul{0.54} / \tul{1.16} & 0.59 / 1.23 & 0.57 / 1.21 & 0.57 / \tul{1.16} \\
ZARA1 & 0.47 / 1.00 & 0.42 / 0.91 & 0.56 / 1.23 & \tul{0.41} / 0.91 & 0.44 / 0.96 & 0.42 / 0.94 & 0.82 / 1.85 & 0.74 / 1.59 & \tbf{0.40} / \tbf{0.87} & 0.44 / 0.97 & 0.42 / \tul{0.89} & 0.45 / 0.99 & 0.51 / 1.01 \\
ZARA2 & 0.56 / 1.17 & 0.52 / 1.11 & 0.44 / 0.90 & \tbf{0.31} / \tbf{0.68} & 0.35 / 0.77 & \tul{0.32} / \tul{0.71} & 0.62 / 1.46 & 0.60 / 1.31 & 0.35 / 0.77 & 0.33 / 0.74 & \tbf{0.31} / \tbf{0.68} & 0.34 / 0.75 & 0.38 / 0.74 \\ \cmidrule(lr){1-14}
AVG   & 0.72 / 1.54 & 0.67 / 1.41 & 0.60 / 1.23 & 0.54 / 1.11 & 0.53 / 1.14 & 0.52 / 1.11 & 0.82 / 1.85 & 0.83 / 1.75 & 0.53 / 1.13 & 0.54 / 1.12 & \tul{0.52} / \tul{1.08} & \tbf{0.51} / 1.11 & \tbf{0.48} / \tbf{0.88} \\
SDD   & 31.2 / 57.0 & 27.3 / 41.4 & 31.4 / 56.8 & 28.0 / 41.3 & 28.8 / 51.4 & 22.7 / 42.0 & 29.8 / 65.1 & 25.2 / 57.6 & 24.7 / 49.0 & 24.2 / 49.3 & 22.1 / \tul{38.0} & \tul{20.7} / 41.9 & \tbf{17.5} / \tbf{34.5} \\
GCS   & 40.2 / 67.2 & 33.6 / 50.5 & 31.9 / 48.4 & 31.8 / 49.3 & 29.3 / 46.5 & 16.9 / 35.1 & 28.3 / 61.2 & 19.4 / 41.5 & 16.7 / \tul{34.9} & \tul{16.6} / 35.0 & \tbf{16.5} / 36.3 & 17.6 / 37.2 & 16.9 / \tbf{34.8} \\
\bottomrule
\\ ~ \vspace{-32pt} ~ \\
\toprule
Stochastic & \tworow{Social-}{GAN\,\cite{gupta2018social}} & \tworow{Social-}{\!\!\!\!S\nsp T\nsp G\nsp C\nsp N\nsp N\cite{mohamed2020social}\!\!\!\!} & \tworow{PECNet$^\dagger$\!}{\cite{mangalam2020pecnet}} & \tworow{Trajectron}{++$^\dagger$\,\cite{salzmann2020trajectron++}} & \tworow{AgentFor}{mer\,\cite{yuan2021agent}} & MID$^\dagger$\,\cite{gu2022mid} & \tworow{GP-Graph}{\cite{bae2022gpgraph}} & NPSN\,\cite{bae2022npsn} & \tworow{\!SocialVAE\!}{\cite{xu2022socialvae}} & \tworow{EqMotion}{\cite{xu2023eqmotion}} & \tworow{EigenTraj}{ectory\,\cite{bae2023eigentrajectory}} & LED\,\cite{mao2023leapfrog} & \tbf{\LBMTPS} \\ \midrule
ETH   & 0.77 / 1.40 & 0.65 / 1.10 & 0.61 / 1.07 & 0.61 / 1.03 & 0.46 / 0.80 & 0.57 / 0.93 & 0.43 / 0.63 & \tbf{0.36} / 0.59 & 0.41 / 0.58 & 0.40 / 0.61 & \tbf{0.36} / \tul{0.53} & \tul{0.39} / 0.58 & 0.41 / \tbf{0.51} \\
HOTEL & 0.43 / 0.88 & 0.50 / 0.86 & 0.22 / 0.39 & 0.20 / 0.28 & 0.14 / 0.22 & 0.21 / 0.33 & 0.18 / 0.30 & 0.16 / 0.25 & 0.13 / 0.19 & \tul{0.12} / 0.18 & \tul{0.12} / 0.19 & \tbf{0.11} / \tul{0.17} & \tul{0.12} / \tbf{0.16} \\
UNIV  & 0.75 / 1.50 & 0.44 / 0.80 & 0.34 / 0.56 & 0.30 / 0.55 & 0.25 / 0.45 & 0.29 / 0.55 & 0.24 / 0.42 & 0.23 / 0.39 & \tbf{0.21} / \tul{0.36} & 0.23 / 0.43 & 0.24 / 0.43 & 0.26 / 0.43 & \tul{0.22} / \tbf{0.34} \\
ZARA1 & 0.35 / 0.69 & 0.34 / 0.53 & 0.25 / 0.45 & 0.24 / 0.41 & 0.18 / 0.30 & 0.28 / 0.50 & 0.17 / 0.31 & 0.18 / 0.32 & \tbf{0.17} / \tbf{0.29} & 0.18 / 0.32 & 0.19 / 0.33 & 0.18 / 0.26 & 0.20 / 0.32 \\
ZARA2 & 0.36 / 0.72 & 0.31 / 0.48 & 0.19 / 0.33 & 0.18 / 0.32 & 0.14 / 0.24 & 0.20 / 0.37 & 0.15 / 0.29 & \tul{0.14} / 0.25 & \tbf{0.13} / \tbf{0.22} & \tbf{0.13} / \tul{0.23} & \tul{0.14} / 0.24 & \tbf{0.13} / \tbf{0.22} & 0.17 / 0.27 \\ \cmidrule(lr){1-14}
AVG   & 0.53 / 1.04 & 0.45 / 0.75 & 0.32 / 0.56 & 0.31 / 0.52 & 0.23 / 0.40 & 0.31 / 0.54 & 0.23 / 0.39 & \tbf{0.21} / 0.36 & \tbf{0.21} / \tul{0.33} & \tbf{0.21} / 0.35 & \tbf{0.21} / 0.34 & \tbf{0.21} / \tul{0.33} & \tul{0.22} / \tbf{0.32} \\
SDD   & 13.6 / 24.6 & 20.8 / 33.2 & 10.0 / 15.9 & 11.4 / 20.1 &~~8.7 / 14.9 &~~7.6 / 14.3 &~~9.1 / 13.8 &~~8.6 / 11.9 &~~8.1 / \tul{11.7} &~~\tul{7.9} / 11.9 &~~8.1 / 13.1 &~~8.5 / \tul{11.7} &~~\tbf{7.8} / \tbf{10.1} \\
GCS   & 15.9 / 32.6 & 14.7 / 23.9 & 17.1 / 29.3 & 12.8 / 24.2 & 10.2 / 16.9 & 10.7 / 18.2 &~~7.8 / 13.7 &~~7.7 / 13.4 &~~\tul{7.4} / \tul{11.9} &~~7.6 / 13.1 &~~\tul{7.4} / 12.5 & N\,/\,A &~~\tbf{7.1} / ~~\tbf{9.6} \\
\bottomrule
\end{tabular}
}
\vspace{-3mm}
\caption{Comparison of \LBMTPS methods with other state-of-the-art deterministic and stochastic methods (ADE\pslp FDE, Unit: meter for ETH\pslp UCY and pixel for SDD\pslp GCS). {\begin{footnotesize}$\dagger$\end{footnotesize}}: Issues raised in the authors' GitHubs are fixed, \textbf{Bold}: Best, \underline{Underline}: Second best.}
\label{tab:result_supervised}
\vspace{-2mm}
\end{table*}

\subsection{Implementation Details}\label{sec:method_implementation}
To demonstrate the zero-shot performance of the proposed \LBMTPZ, we use GPT-3.5 and GPT-4 as foundational language models for prompt engineering. In this experiment, since the proposed model does not require a training procedure, we exclude the tokenizer optimization and multi-task training methods. 
Each API call for one trajectory inference takes about 20 seconds, so we carry out a multi-process by creating a thread pool of 1,000 units to evaluate all paths in the datasets. 
We ensure prediction fidelity from the output sentence by retrying if responses are not aligned with the desired answer format.

For the supervised training, we leverage the full potential of our \LBMTPS model by integrating all proposed techniques in~\cref{sec:method_prompt_conversion,sec:method_domain_shift}. We use the BPE model~\cite{sennrich2016neural} for the tokenizer, and encoder-decoder, T5~\cite{raffel2020t5}, as our backbone language model. The two models are trained using all the multi-task QA templates. T5 is trained using a cross-entropy loss between the generated outputs and the tokenized ground-truth answers in an end-to-end manner. The hyper-parameters $d$ and $\tau$ in~\cref{sec:method_domain_shift} are empirically set to 2 and 0.7, respectively. AdamW optimizer~\cite{loshchilov2018decoupled} is used, whose batch size is 512 and learning rate is 1e-4 over 200 epochs. The training time takes about 4 hours, leveraging a distributed data parallel pipeline on a machine of 8 NVIDIA 4090 GPUs.

\section{Experiments}\label{sec:experiment}
In this section, we conduct comprehensive experiments to verify the effectiveness of our language-based approach for trajectory prediction. We first describe the experimental setup in~\cref{sec:experiment_setup}. We then provide comparison results with various zero-shot and supervised trajectory prediction methods in~\cref{sec:experiment_result}. We lastly conduct an extensive ablation study to validate the effect of each component in our method in \cref{sec:experiment_ablation}.

\subsection{Experimental Setup}\label{sec:experiment_setup}
\vspace{0mm}\noindent\textbf{Datasets.}\quad
We conduct experiments on four public datasets: ETH~\cite{pellegrini2009you}, UCY~\cite{lerner2007crowdsbyexample}, Stanford Drone Dataset (SDD)~\cite{robicquet2016learning}, and the Grand Central Station (GCS)~\cite{yi2015understanding} dataset to compare our \LBMTP model with state-of-the-art trajectory predictors. The ETH and UCY datasets consist of five subset scenes (ETH, Hotel, Univ, Zara1 and Zara2) with 1,536 pedestrians recorded with the surveillance camera. We use the standard train-val-test split and adopt the leave-one-out strategy~\cite{alahi2016social,gupta2018social} for the training and evaluation. SDD has 5,232 trajectories of six agent categories, including pedestrians, cars, and bicyclists, in eight different university campus scenes from a top-down drone view. GCS shows a highly congested terminal scene with 12,684 pedestrians streaming to the exit. We follow the standard benchmark protocol in~\cite{gupta2018social,huang2019stgat,mohamed2020social,Shi2021sgcn,bae2022npsn} that the first 3.2 seconds ($T_{\textit{obs}}\!=\!8$ frames) are used as observations, and the following 4.8 seconds ($T_{\textit{pred}}\!=\!12$ frames) are predicted.

\vspace{0.5mm}\noindent\textbf{Evaluation protocols.}\quad
To evaluate the tokenizer for \LBMTP, we use the Recall-Oriented Understudy for Gisting Evaluation (ROUGE) score from the NLP task to measure the text similarity. Specifically, ROUGE-1 checks the overlap ratio of each word between the source and target sentence.
In order to measure the prediction accuracy of \LBMTP, we use two metrics as accuracy measures: Average Displacement Error (ADE) and Final Displacement Error (FDE). The ADE and FDE compute the Euclidean distance between a predicted and a ground-truth path and their destination, respectively. Following~\cite{gupta2018social}, we generate $K\!=\!1$ samples for the most-likely evaluation and $K\!=\!20$ samples for the multimodal trajectory prediction. For the multimodal predictions, we generate multiple outputs and then choose the best path for the performance evaluation.

\begin{figure*}[t]
\centering
\includegraphics[width=\linewidth,trim={10mm 0 10mm 0},clip]{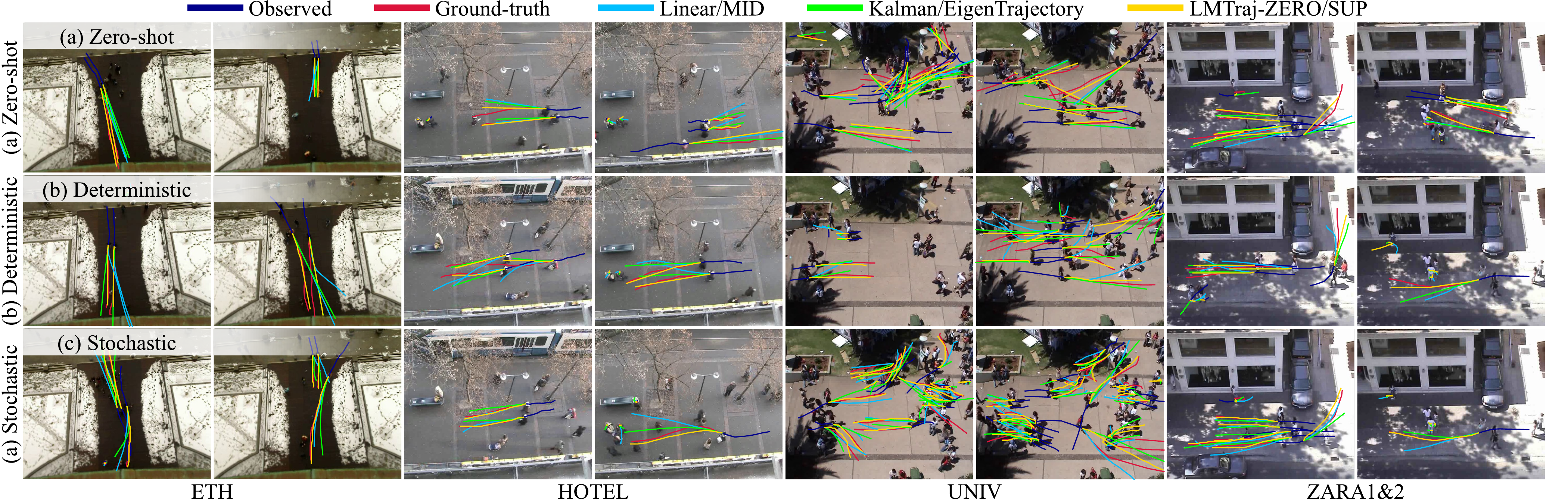}
\vspace{-7mm}
\caption{Visualization of prediction results on (a) zero-shot and two supervised trajectory prediction benchmarks: (b) deterministic and (c) stochastic approach. To aid visualization for the stochastic approach, we report one best trajectory of $K=20$ samples each.}
\label{fig:result_visualize}
\end{figure*}

\subsection{Evaluation Results}\label{sec:experiment_result}
\vspace{0mm}\noindent\textbf{Evaluation of the numerical tokenizer.}\quad
First, we check the efficiency of our numerical tokenizers compared to pretrained tokenizers in~\cite{raffel2020t5} trained on texts. To find the most suitable tokenizer type for numerical trajectory data, we test four types: char, word, unigram, and byte pair encoding (BPE) using six forecasting QA prompts. In particular, the Char-based model~\cite{Sutskever2011Generating} breaks the text down into individual characters, while the word-based model~\cite{pennington2014glove} splits the text into words, which are separated by whitespace. BPE~\cite{sennrich2016neural} tokenizes sentences by iteratively searching the text and by repeatedly merging the most frequent sequence pairs of letters in a vocabulary. The Unigram model~\cite{kudo2018subword} uses an approach similar way to BPE, but generates a vocabulary by lexicalizing byte pairs based on probability values for neighboring characters. Additionally, the pretrained model employs the unigram model trained on 750GB of web crawled text data~\cite{raffel2020t5}.

\Cref{tab:result_tokenizer} shows that all five methods cover all the words in the trajectory prompts well. Since no vocabulary is missing, the input and output sentences are exactly the same as the original sentence after tokenization. However, the pretrained and word tokenizers often have a mixture of letters and numbers. As shown in \cref{fig:comparison_tokenizer}-(b,d), certain tokens combine letters and numbers, and even multiple tokens are used to represent the integer part of the number 6. This disturbs \LBMTP's understanding of the sequential nature of the numbers. While a Char tokenizer is capable of separating numerical and letter notation, it requires too many tokens for tokenization, as in \cref{fig:comparison_tokenizer}-(c). Both unigram and BPE tokenizers effectively distinguish between letters and numbers while decreasing the average number of tokens by merging multi-digit numbers into a single token in \cref{fig:comparison_tokenizer}-(e,f). This directly reduces the computational complexity of \LBMTP. We select the BPE tokenizer for our \LBMTPS model, thanks to its ability to represent a sentence with a smaller number of tokens.

\vspace{0.5mm}\noindent\textbf{Evaluation of the zero-shot approach.}\quad
To demonstrate the potential of prompt engineering with language foundation models for trajectory prediction, we conduct a quantitative comparison between \LBMTPZ and  various zero-shot methods. We provide three algorithmic approaches and one learnable model. The `Stop' operates by stopping walking at the final observation point without making predictions, while `Linear' and `Kalman filter' methods serve as state extrapolation techniques. AutoTrajectory~\cite{ma2020autotrajectory} is trained in an unsupervised manner without any ground-truth trajectory label. As shown in~\cref{tab:result_zeroshot} and \cref{fig:result_visualize}-(a), we observe that our method achieves the best performance among all the zero-shot methods. Particularly, using \LBMTPZ with GPT-4 yields results superior to that of GPT-3.5, indicating the model has better performance when combined with larger language foundation models. \LBMTPZ with GPT-4 shows comparable performance to the supervised model, Social-STGCNN~\cite{mohamed2020social}. This opens the possible study of zero-shot trajectory prediction.

\begin{figure}[t]
\centering
\includegraphics[width=\linewidth,trim={0 106mm 0 0},clip]{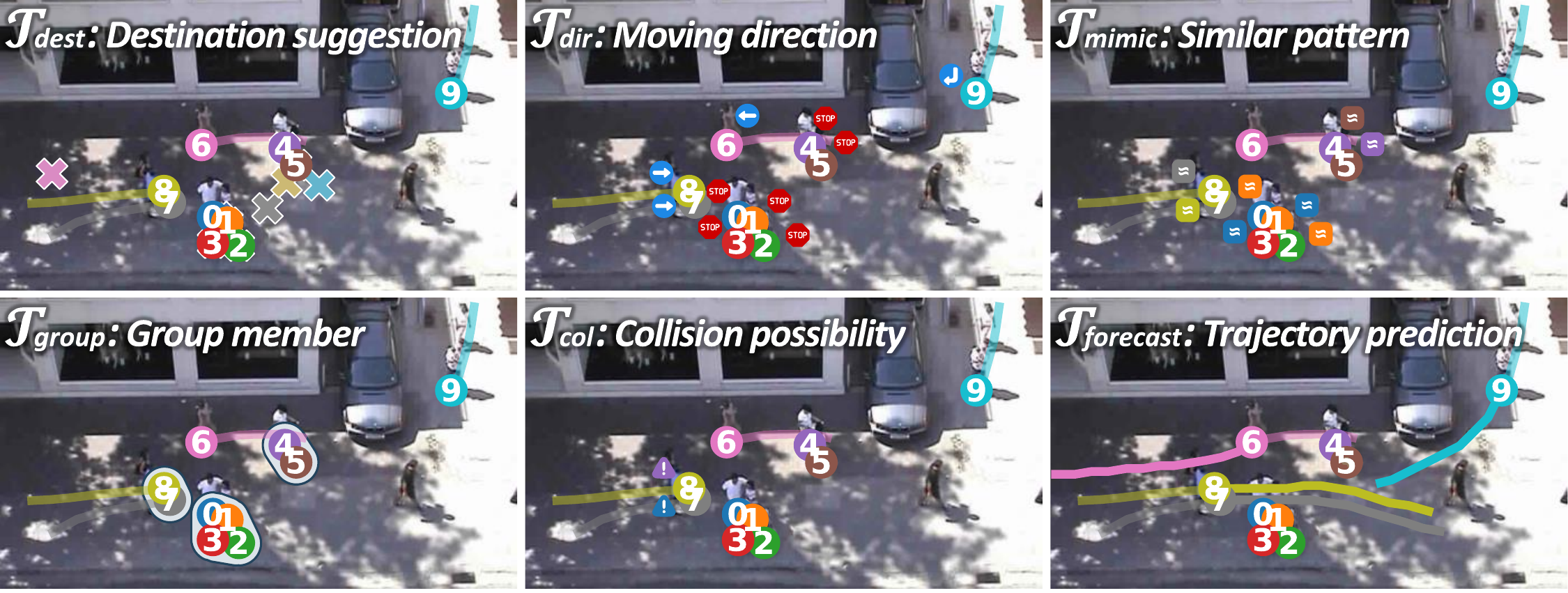}
\vspace{-7mm}
\caption{Visualization of the social reasoning using observed paths and the corresponding trajectory prediction results. }
\label{fig:result_socialreasoning}
\end{figure}

\vspace{0.5mm}\noindent\textbf{Evaluation of the supervised approach.}\quad
Next, to check the maximum performance of the linguistic approach, we compare \LBMTPS to both deterministic and stochastic trajectory prediction methods. As shown in~\cref{tab:result_supervised}, \LBMTPS outperforms the deterministic predictions on most datasets, while the other models reached a plateau. This demonstrates the effectiveness of the \LBMTPS for reasoning about complex social relationships in~\cref{fig:result_socialreasoning} as well as performing beam search based on cumulative probabilities for the most likely path, as visualized in~\cref{fig:result_visualize}-(b). This provides a significant advantage over the previous works that rely on the graph-based social interaction modeling and the greedy selection of footsteps.

In addition, \LBMTPS also shows promising results for stochastic trajectory prediction. By understanding potential future behavior patterns through scene descriptions and social reasoning, \LBMTPS, generating sentences of realistic trajectories, achieves better performance than the previous works. As shown in~\cref{fig:result_visualize}-(c), our \LBMTPS successfully generates multimodal trajectories using the temperature tuning technique to diversify the outputs, as in NLP. This means that our approach offers a new potential solution to the limited performance of existing physics-based social relationships.

\subsection{Ablation Studies}\label{sec:experiment_ablation}
\vspace{0mm}\noindent\textbf{Effectiveness of the numerical tokenizer.}\quad
We compare the effectiveness of the text-based pretrained tokenizer with our numerical tokenizer for stochastic trajectory prediction. In~\cref{tab:ablation_component}, our \LBMTPS with the numerical tokenizer outperforms the pretrained tokenizer in deterministic prediction accuracy. This shows the advantage of our approach for numerical tasks, by allowing the model to better understand numerical information from sentences.

\begin{table}[t]
\Large
\centering
\resizebox{\linewidth}{!}{%
\begin{tabular}{@{}c@{}ccccccc}
    \toprule
    \multicolumn{2}{c}{Model} & ETH & HOTEL & UNIV & ZARA1 & ZARA2 & AVG \\ \midrule
    \multirow{2}{*}{Tokenizer}   & \!Pretrained\! & 0.85\pslp1.49 & 0.46\pslp0.93 & 0.97\pslp2.00 & 0.55\pslp1.06 & 0.43\pslp0.89 & 0.65\pslp1.28 \\
                                 & \!Numerical\!  & \tbf{0.65}\pslp\tbf{1.04} & \tbf{0.26}\pslp\tbf{0.46} & \tbf{0.57}\pslp\tbf{1.16} & \tbf{0.51}\pslp\tbf{1.01} & \tbf{0.38}\pslp\tbf{0.74} & \tbf{0.48}\pslp\tbf{0.88} \\ \cmidrule(lr){1-8}
    \multirow{3}{*}{Size}        & Small          & \tbf{0.65}\pslp\tbf{1.04} & 0.26\pslp\tul{0.46} & 0.57\pslp1.16 & \tul{0.51}\pslp\tul{1.01} & \tbf{0.38}\pslp\tul{0.74} & 0.48\pslp\tbf{0.88} \\
                                 & Medium         & \tul{0.68}\pslp\tul{1.17} & 0.26\pslp\tbf{0.45} & 0.57\pslp1.16 & \tul{0.51}\pslp1.02 & \tul{0.39}\pslp0.76 & 0.48\pslp\tul{0.91} \\
                                 & Large          & 0.71\pslp1.22 & 0.26\pslp\tul{0.46} & 0.57\pslp1.16 & \tbf{0.50}\pslp\tbf{1.00} & \tbf{0.38}\pslp\tbf{0.73} & 0.48\pslp\tul{0.91} \\ \cmidrule(lr){1-8}
    \multirow{2}{*}{Multi-task}  & No         & 0.74\pslp1.27 & 0.31\pslp0.59 & 0.74\pslp1.51 & 0.53\pslp1.06 & 0.41\pslp0.79 & 0.55\pslp1.04 \\ 
                                 & Yes         & \tbf{0.66}\pslp\tbf{1.07} & 0.26\pslp0.46 & 0.57\pslp\tbf{1.16} & \tbf{0.52}\pslp\tbf{1.02} & 0.38\pslp\tbf{0.74} & 0.48\pslp\tbf{0.89} \\ \cmidrule(lr){1-8}
    \multirow{5}{*}{Depth}       & $d\!=\!1$      & \tul{0.66}\pslp\tul{1.05} & 0.26\pslp0.46 & 0.57\pslp\tul{1.17} & \tbf{0.51}\pslp\tbf{1.00} & 0.38\pslp0.75 & 0.48\pslp\tul{0.89} \\ 
                                 & $d\!=\!2$      & \tbf{0.65}\pslp\tbf{1.04} & 0.26\pslp0.46 & 0.57\pslp\tbf{1.16} & \tbf{0.51}\pslp\tul{1.01} & 0.38\pslp\tul{0.74} & 0.48\pslp\tbf{0.88} \\ 
                                 & $d\!=\!3$      & \tul{0.66}\pslp1.07 & 0.26\pslp0.46 & 0.57\pslp\tbf{1.16} & \tul{0.52}\pslp1.02 & 0.38\pslp\tul{0.74} & 0.48\pslp\tul{0.89} \\ 
                                 & $d\!=\!4$      & 0.67\pslp1.09 & 0.26\pslp0.46 & 0.57\pslp\tbf{1.16} & \tul{0.52}\pslp1.03 & 0.38\pslp\tbf{0.73} & 0.48\pslp\tul{0.89} \\ 
                                 & $d\!=\!5$      & 0.67\pslp1.10 & 0.26\pslp0.46 & 0.57\pslp\tbf{1.16} & \tul{0.52}\pslp1.03 & 0.38\pslp\tul{0.74} & 0.48\pslp0.90 \\ \cmidrule(lr){1-8}
    \multirow{5}{*}{Temperature} & $\tau\!=\!0.1$ & 0.49\pslp0.70 & 0.19\pslp0.31 & 0.41\pslp0.80 & 0.34\pslp0.64 & 0.29\pslp0.53 & 0.34\pslp0.60 \\ 
                                 & $\tau\!=\!0.3$ & 0.45\pslp0.58 & 0.15\pslp0.21 & 0.29\pslp0.52 & 0.26\pslp0.45 & 0.22\pslp0.38 & 0.27\pslp0.43 \\ 
                                 & $\tau\!=\!0.5$ & \tul{0.42}\pslp0.54 & \tul{0.13}\pslp\tul{0.17} & \tul{0.24}\pslp0.41 & \tul{0.21}\pslp0.36 & 0.19\pslp\tul{0.31} & 0.24\pslp0.36 \\ 
                                 & $\tau\!=\!0.7$ & \tbf{0.41}\pslp\tbf{0.51} & \tbf{0.12}\pslp\tbf{0.16} & \tbf{0.22}\pslp\tbf{0.34} & \tbf{0.20}\pslp\tbf{0.32} & \tbf{0.17}\pslp\tbf{0.27} & \tbf{0.22}\pslp\tbf{0.32} \\ 
                                 & $\tau\!=\!0.9$ & \tul{0.42}\pslp\tul{0.53} & \tul{0.13}\pslp0.18 & \tbf{0.22}\pslp\tul{0.35} & 0.22\pslp\tul{0.35} & \tul{0.18}\pslp\tbf{0.27} & \tul{0.23}\pslp\tul{0.34} \\
    \bottomrule
\end{tabular}
}
\vspace{-3mm}
\caption{Ablation studies on each component of \LBMTPS (ADE/FDE, meter). \tbf{Bold}: Best, \tul{Underline}: Second best.}
\label{tab:ablation_component}
\end{table}

\vspace{0.5mm}\noindent\textbf{Model size.}\quad
Next, we vary the sizes of the backbone Seq2Seq model in~\cref{tab:ablation_component}. As expected, the performance varies slightly with increasing model size, but the gain is marginal. As a result, we choose the smallest and the light-weight model for the real-time prediction.

\vspace{0.5mm}\noindent\textbf{Multi-task training strategy.}\quad
To enhance the model's ability to reason about social interactions, we include a multitask training in the forecasting pipeline. As reported in~\cref{tab:ablation_component}, the multi-task training strategy improves the performance compared to a single-task training strategy. This improvement suggests that integrating domain knowledge pushes the model to better understand group behaviors and collision avoidance, helpful for the main forecasting task.

\begin{figure}[t]
\centering
\includegraphics[width=\linewidth,trim={0 116mm 0 0},clip]{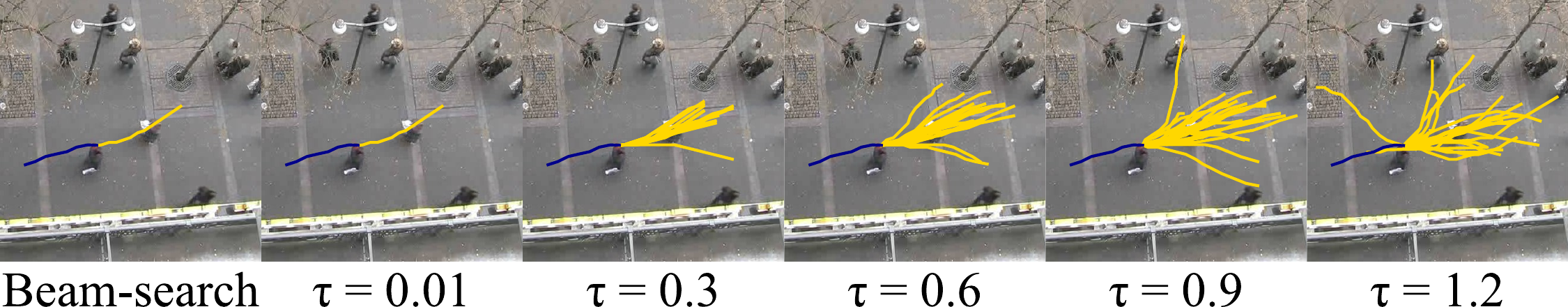}
\vspace{-6mm}
\caption{Visualization of the most-likely and multimodal trajectory generation capability of our \LBMTPS ($\tau$: temperature).}
\label{fig:result_temperature}
\end{figure}

\vspace{0.5mm}\noindent\textbf{Beam-search and temperature analysis.}\quad
We conduct a parameter study for both the deterministic and stochastic predictions. \Cref{tab:ablation_component} validates that using beam search with a depth of $d\!=\!2$ and the temperature-tuning with $\tau\!=\!0.7$ produces the best performance. In addition, adjusting the temperature parameter $\tau$ affects the level of uncertainty in the multi-modal generation, allowing for controlled variations within socially acceptable limits in~\cref{fig:result_temperature}

\vspace{0.5mm}\noindent\textbf{Computational cost.}\quad
Lastly, we check the computational efficiency of \LBMTPS in \cref{tab:ablation_efficiency}. Due to the structural nature of the language model that sequentially predicts the next token, the inference time is a little slower than the fastest model. However, it produces promising results with the reasonable GPU memory consumption as well as real-time inference.

\begin{table}[t]
\large
\fontsize{12}{13.5}\selectfont
\centering
\resizebox{\linewidth}{!}{%
\begin{tabular}{ccccccc}
\toprule
\multirow{2}{*}{Model\vspace{-5pt}} & \multicolumn{2}{c}{Accuracy} & & \multicolumn{3}{c}{Complexity} \\ \cmidrule(lr){2-3} \cmidrule(lr){5-7}
             & ~ADE~  & ~FDE~ & & ~GPU\,memory & Training & Inference~ \\ \midrule
PECNet~\cite{mangalam2020pecnet}             & 0.32 & 0.56 & & ~\tul{1733\,MB}  & ~~\tbf{0.3\,h}  & 57.0\,ms~  \\
MID~\cite{gu2022mid}                         & 0.31 & 0.54 & & ~2929\,MB  & ~~6.9\,h & 35.0\,ms~  \\
STAR~\cite{yu2020spatio}                     & 0.26 & 0.53 & & ~1735\,MB  & 36.3\,h  & 97.0\,ms~  \\
~~~AgentFormer~\cite{yuan2021agent}~~~       & 0.23 & 0.40 & & ~9639\,MB  & 22.0\,h  & ~~\tbf{8.2\,ms}~ \\
SocialVAE~\cite{xu2022socialvae}             & \tbf{0.21} & \tul{0.33} & & ~1762\,MB & ~~\tul{2.1\,h} & 73.0\,ms~ \\ \cmidrule(lr){1-7}
\tbf{\LBMTPS}                                & \tul{0.22} & \tbf{0.32} & & ~\tbf{1401\,MB}  & ~~3.8\,h & \tul{18.3\,ms}~  \\
\bottomrule
\end{tabular}
}
\vspace{-3mm}
\caption{Computational complexity analysis of our \LBMTPS with other numerical-based trajectory prediction models. `Inference' measures the average inference time per trajectory.}
\label{tab:ablation_efficiency}
\vspace{-1mm}
\end{table}

\section{Conclusion}
This paper demonstrates the ability of language models to understand and extrapolate spatio-temporal numeric information from trajectory data. We shift the domain of the trajectory prediction task to a question-answering task, which provides historical data as context and then forecasts futures when answering the given question templates. The history data, transformed into a text prompt format, can offer rich information for the language model, and capture human dynamics. We show that both the prompt engineering of the language foundation models and their end-to-end training can successfully predict accurate future paths in zero-shot and supervised manners using our \LBMTPZ and \LBMTPS, respectively. In addition, the specialized techniques for large language models, including tokenizer optimization, multi-task learning, beam-search, and temperature tuning scheme, allow our model to better comprehend high-level social reasoning, and to operate like conventional deterministic and stochastic trajectory predictor models.

\vspace{1mm}
\fontsize{8}{10}\selectfont{%
\noindent\textbf{Acknowledgement}
This research was supported by `Project for Science and Technology Opens the Future of the Region' program through the INNOPOLIS FOUNDATION funded by Ministry of Science and ICT (Project Number: 2022-DD-UP-0312), Vehicles AI Convergence Research $\&$ Development Program through the National IT Industry Promotion Agency of Korea (NIPA) funded by the Ministry of Science and ICT (No.S1602-20-1001), and the Institute of Information $\&$ communications Technology Planning $\&$ Evaluation (IITP) grant funded by the Korea government (MSIT) (No.2019-0-01842, Artificial Intelligence Graduate School Program (GIST) and No.2021-0-02068, Artificial Intelligence Innovation Hub).}

{
    \small
    \bibliographystyle{ieeenat_fullname}
    \bibliography{egbib}
}

\end{document}